\documentclass{article}

\usepackage{subfigure}

\usepackage{hyperref}

\usepackage[accepted]{icml2025}

\usepackage{algorithmic}

\usepackage{mathtools} 
\usepackage{amssymb}   
\usepackage{amsthm}    
\usepackage{bbm}       
\usepackage{bm}        

\theoremstyle{plain}

\theoremstyle{definition}

\theoremstyle{remark}

\usepackage[T1]{fontenc}
\usepackage[utf8]{inputenc}
\usepackage[nopatch=footnote]{microtype}  
\usepackage{times}                        
\usepackage{inconsolata}                  
\usepackage[scaled=0.85]{helvet}          
\usepackage{graphicx}                     
\makeatletter                             
    \DeclareRobustCommand*{\escapeus}[1]{%
    \begingroup\@activeus\scantokens{#1\endinput}\endgroup}
    \begingroup\lccode`\~=`\_\relax
   \lowercase{\endgroup\def\@activeus{\catcode`\_=\active \let~\_}}
\makeatother
\newcommand{\myemph}[1]{\textsf{{\escapeus{#1}}}}

\usepackage{enumitem}        
\usepackage{fnpct}           
\usepackage{tcolorbox}       
\usepackage[misc]{ifsym}     
\usepackage{url}             
\usepackage{booktabs}        
\usepackage{makecell}        
\usepackage{multirow}        
\usepackage{csquotes}        
\usepackage{xspace}          
\makeatletter                
\xspaceaddexceptions{\csq@qclose@i}
\makeatletter

\usepackage[table-column-type=s]{siunitx}
\DeclareSIUnit\thousand{k}
\DeclareSIUnit\million{M}
\DeclareSIUnit\billion{B}
\DeclareSIUnit\trillion{T}
\DeclareSIUnit\x{x}
\DeclareSIUnit\percent{\%}
\DeclareSIUnit\hour{h}
\DeclareSIUnit\min{m}
\DeclareSIUnit\sec{s}
\DeclareSIUnit\gb{GB}
\newcommand{\integer}[1]{\num[mode = math, round-mode=places, round-precision=0, group-separator={,}, group-minimum-digits=4]{#1}\xspace}
\newcommand{\float}[2][1]{\num[group-digits=false, round-precision=#1, round-mode=places]{#2}\xspace}
\newcommand{\snum}[1]{\num{#1}\xspace}
\newcommand{\q}[2]{\qty[mode=math]{#1}{#2}\xspace}

\usepackage{tabularray}
\usepackage{cellspace}
\usepackage{setspace}
\usepackage[capitalize,noabbrev]{cleveref} 
\usepackage{cleveref}
\crefname{section}{\S}{\S\S}
\Crefname{section}{\S}{\S\S}
\crefformat{section}{\S#2#1#3}
\crefname{appendix}{App.}{}
\crefname{figure}{Fig.}{Fig.}
\crefname{table}{Table}{Tables}
\crefname{equation}{eq.}{eqs.}
\crefformat{equation}{eq.~#2#1#3}
\crefrangeformat{equation}{eqs.~#3#1#4 to #5#2#6}
\crefmultiformat{equation}{eqs.~#2#1#3}{ and~#2#1#3}{, #2#1#3}{ and~#2#1#3}

\usepackage[textsize=tiny]{todonotes}

\newcommand{\example}[1]{\textit{\enquote{#1}}}

\newcommand{\defn}[1]{\textbf{#1}}
\newcommand{\vs}{\emph{vs.}\@\xspace}

\newcommand{\bpe}{\myemph{BPE}\xspace}
\newcommand{\bpewp}{\myemph{BPE-WP}\xspace}
\newcommand{\wordpiece}{\myemph{WP}\xspace}

\newcommand{\ngram}{$n$-gram\xspace}
\newcommand{\ngrams}{$n$-grams\xspace}
\newcommand{\fineweb}{\myemph{FineWeb-Edu}\xspace}
\newcommand{\commoncorpus}{\myemph{Common-Corpus}\xspace}
\newcommand{\blimp}{\myemph{BLiMP}\xspace}
\newcommand{\minipile}{\myemph{MiniPile}\xspace}

\newcommand{\renyi}{R\'enyi\xspace}

\newcommand{\ch}{b}
\newcommand{\chvec}{\mathbf{\ch}}

\newcommand{\dataset}{\mathcal{D}}

\newcommand{\tokname}{\myemph{ByteSpan}\xspace}
\newcommand{\thres}{\theta}

\definecolor{customgreen}{HTML}{3BAE2E}  
\definecolor{customred}{HTML}{D13438}  
\definecolor{customyellow}{HTML}{E0A000} 
\newcommand{\green}[1]{\textcolor{customgreen}{#1}}
\newcommand{\red}[1]{\textcolor{customred}{#1}}
\newcommand{\yellow}[1]{\textcolor{customyellow}{#1}}

\icmltitlerunning{\texorpdfstring{\tokname}{ByteSpan}: Information-Driven Subword Tokenisation}

\begin{document}

\twocolumn[
\icmltitle{\texorpdfstring{\tokname}{ByteSpan}: Information-Driven Subword Tokenisation}

\icmlsetsymbol{equal}{*}

\begin{icmlauthorlist}
    \icmlauthor{Z\'ebulon Goriely}{cam}
    \icmlauthor{Suchir Salhan}{equal,cam,alta}
    \icmlauthor{Pietro Lesci}{equal,cam}
    \icmlauthor{Julius Cheng}{cam}
    \icmlauthor{Paula Buttery}{cam,alta}
\end{icmlauthorlist}

\icmlaffiliation{cam}{Department of Computer Science and Technology, University of Cambridge, U.K.}
\icmlaffiliation{alta}{ALTA Institute, University of Cambridge, U.K.}

\icmlcorrespondingauthor{Z\'ebulon Goriely}{zebulon.goriely@cl.cam.ac.uk}

\icmlkeywords{Morphology, Tokenization, Tokenisation, Subwords, ICML}

\vskip 0.3in
]



\printAffiliationsAndNotice{\icmlEqualContribution} 

\begin{abstract}
    Recent dynamic tokenisation methods operate directly on bytes and pool their latent representations into \textit{patches}. This bears similarities to computational models of word segmentation that determine lexical boundaries using spikes in an autoregressive model's prediction error. Inspired by this connection, we explore whether grouping predictable bytes---rather than pooling their representations---can yield a useful fixed subword vocabulary. We propose a new information-driven subword tokeniser, \tokname, that uses an external byte-level LM during training to identify contiguous predictable byte sequences and group them into subwords. Experiments show that \tokname yields efficient vocabularies with higher morphological alignment scores than \bpe for English. Multilingual experiments show similar compression and \renyi efficiency for \integer{25} languages. 
\end{abstract}

\section{Introduction}
\label{sec:intro}

Modern language models (LMs) process text as sequences of \textit{byte\footnote{We use \textit{bytes} and \textit{characters} interchangeably.} spans}, or \defn{subwords}, to improve computational efficiency \citep{zouhar-etal-2023-tokenization}.  Processing raw bytes leads to longer sequences, while operating on words requires a large vocabulary.  Subword tokenisation---i.e., grouping bytes into subwords drawn from a fixed, finite vocabulary---offers a balance but is sensitive to spelling \citep{chai-etal-2024-tokenization} and has inconsistent compression rates across languages \citep{rust-etal-2021}. 

To remove the LMs' dependence on tokenisers while preserving computational efficiency, recent work on tokenisation proposes to operate directly on bytes and pool their representations into \textit{patches}. These patches are created either by pooling fixed-length spans (\citealp{dai-etal-2020-funnel, nawrot-etal-2022-hierarchical, yu2023megabyte}, \textit{inter alia}) or by dynamically pooling predictable byte sequences within a context window \citep{nawrot-etal-2023-efficient, pagnoni2024byte}.

Dynamic patching relies on trainable model components and unwittingly mirrors work in child language acquisition, where computational models---ranging from \ngrams to Transformers \citep{vaswani-etal-2017-attention}---are used to study how children may use distributional statistics to group predictable sequences of \textit{phonemes} into words. As with dynamic patching, these methods draw on the simple principle that predictability within lexical units is high, and predictability between lexical units is low \citep{harris1955phoneme}. Inspired by this connection, we explore whether grouping predictable bytes---rather than pooling their representations---can yield a useful fixed subword vocabulary.

We propose a new information-driven tokeniser, \tokname, which uses an external byte-level LM to identify predictable contiguous byte sequences, using entropy or surprisal as measures of information. Byte spans are identified using a global threshold constraint, a monotonic constraint, or a combination of the two, and we propose several methods for using byte spans identified in a training corpus to create a subword vocabulary. We use these methods to train English and multilingual tokenisers and compare them to Byte-Pair Encoding (\bpe, \citealp{sennrich-etal-2016-neural}) across vocabulary sizes. Intrinsic results show that our method yields higher morphological alignment scores and higher \renyi efficiency scores for most vocabulary sizes, without compromising compression. Our multilingual experiments show similar \renyi efficiency and fertility to \bpe across 25 languages and we propose methods for balancing a vocabulary to efficiently tokenise rare orthographies. 

\section{\texorpdfstring{\tokname}{ByteSpan} Tokenisation}
\label{sec:methodology}

Our method, \tokname, uses an external byte-level LM during training to identify predictable byte sequences and group them into subwords. During inference, the resulting vocabulary can be paired with any standard tokenisation function (e.g., longest-prefix match) and modern LM (e.g., \citealp{touvron2023llama}) without modifications. 

\begin{figure*}[!t]
    \centering
    \includegraphics[width=0.8\linewidth]{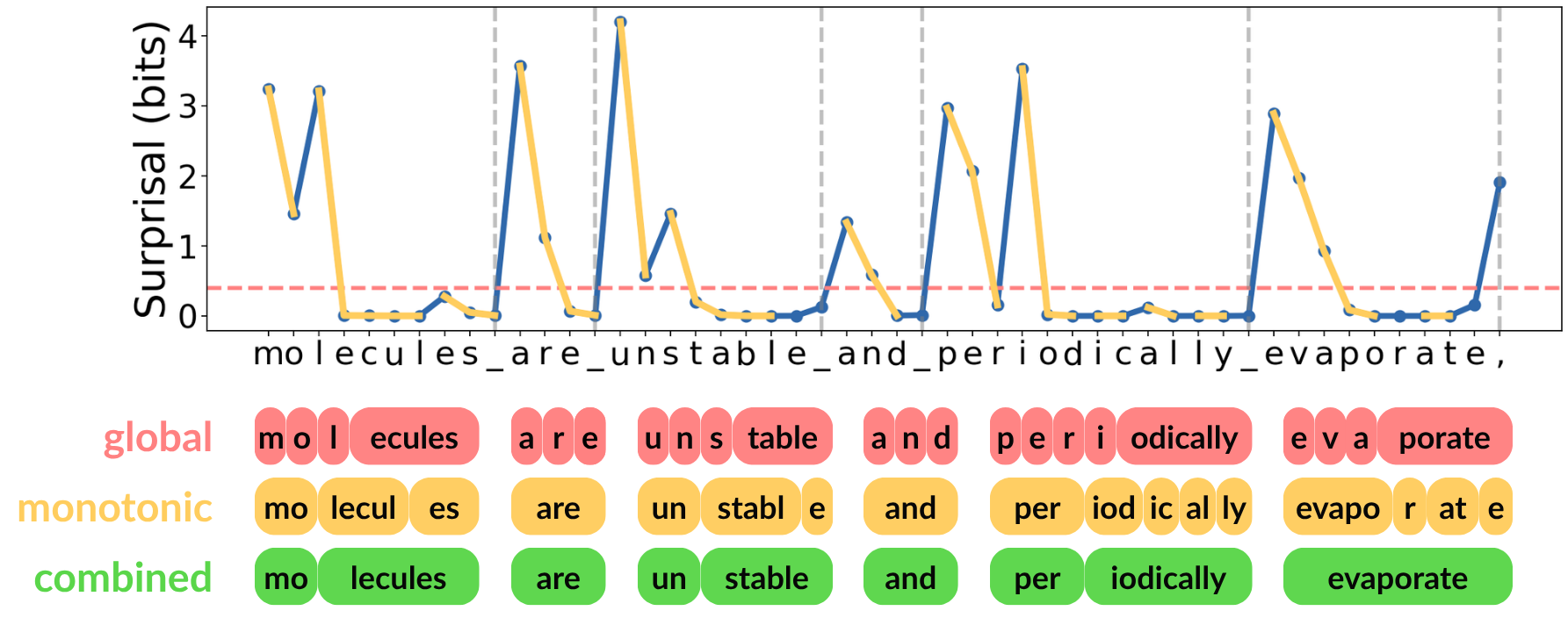}
    \caption{\textbf{Information-Driven Subword Creation.}
    Per-byte \textbf{surprisal} of \textit{\enquote{molecules are unstable and periodically evaporate}} from a byte-level LM. \tokname groups contiguous bytes using one of three constraints; the \red{global constraint} uses a fixed threshold, the \yellow{monotonic constraint} groups bytes with decreasing information and the \green{combined constraint} groups bytes that meet either constraint. Grey vertical lines indicate pre-tokenisation boundaries.
    }
    \label{fig:example}
\end{figure*}

\subsection{Motivation}

This method is related to the work of \citet{pagnoni2024byte}, who group bytes into dynamic `patches' according to their entropies, as given by a byte-level LM. They experiment with two methods to identify patch boundaries; their \textbf{global constraint} identifies bytes whose entropies exceed a fixed threshold and their \textbf{monotonic constraint} identifies bytes whose entropies decrease monotonically. A \textbf{patching function} then segments a stream of bytes into patches that are fed through a latent transformer, with predicted patches decoded by the smaller byte-level LM. Conceptually, this smaller LM can make the relatively `easy' next-byte predictions given the low entropy of each byte within the patch. 


We draw parallels between these patching constraints and computational models of word segmentation. These models are designed to demonstrate how distributional information could be leveraged by language-learning infants to bootstrap a vocabulary, following the influential statistical learning experiments of \citet{Saffran1996learning} who observed this ability in young infants. In the typical framework, these models use unsupervised algorithms to group unsegmented sequences of phonemes from transcriptions of child-directed speech into word-like units. One approach involves maximising the likelihood of word-level \ngram models(e.g., \citealp{Brent1999, Venkataraman2001}), a method that closely resembles the UnigramLM tokenisation algorithm \citep{kudo-2018-subword}. Another approach is to extract measures of uncertainty using phoneme-level \ngram models and posit boundaries where these measures spike or surpass a threshold (e.g., \citealp{ccoltekin2014explicit, goriely2023word}). Neural language models have also been used, for instance by using the prediction of an utterance boundary (which is included in the phoneme sequence) to posit a word boundary \citep{christiansen1998learning}. In a formative analysis of character-level RNNs, \citet{elman1990finding} noted that the prediction error from a neural language model could serve as a cue for lexical boundaries more broadly; boundaries around words, morphemes, but also frequent multi-word sequences, which children often treat as fixed lexical items \citep{macwhinney1978acquisition}. Based on this observation, \citet{goriely2025babylm} trained phoneme-level GPT-2 LMs across 31 languages and demonstrated a method for extracting word boundaries from the trained models; computing model uncertainty using entropy, rank and surprisal from the predictions at each point and segmenting at points of high uncertainty. These information-based approaches to word segmentation, and the later study in particular, closely resemble the patching constraints of \citet{pagnoni2024byte}.

We hypothesise that a modular tokenisation pipeline using constraints based on byte-level information might retain the conceptual advantages of patch-based approaches while retaining the benefits of creating a fixed vocabulary as a pre-processing step before training. By drawing parallels with computational models of word segmentation, we hypothesise that the resulting subword tokens align better with morphological segmentations of natural language than compression-based methods such as \bpe.

\subsection{The \texorpdfstring{\tokname}{ByteSpan} Algorithm}

\tokname works by first collecting predictions from a byte-level LM over a corpus, from which key statistics are calculated, then grouping contiguous sequences of bytes using a constraint based on these statistics.

\paragraph{Statistics and constraints.} As a starting point, we follow the patching methods of \citet{pagnoni2024byte} by using per-byte \textbf{entropy}\footnote{We note that \citet{pagnoni2024byte} actually use \emph{next-byte} entropy, which would shift our byte spans by one unit.} \(H(\ch_i)\) and considering two constraints: 
\begin{align}
    \mathrm{\red{Global~Constraint}} &~~ H(\ch_t) < \thres_g \\
    \mathrm{\yellow{Monotonic~Constraint}} &~~ H(\ch_t) - H(\ch_{t-1}) < 0 
\end{align}

The first constraint groups bytes that fall under a fixed global threshold and the second constraint groups byte with monotonically decreasing information. These are visualised in \cref{fig:example}, where for instance the \red{global constraint} segments the word \example{unstable} as $\{\example{u}, \example{n}, \example{s}, \example{table} \}$ whereas the \yellow{monotonic constraint} produces $\{\example{un}, \example{stabl}, \example{e}\}$. The algorithm for segmenting a sequence using the \yellow{monotonic constraint} is provided in \textit{Algorithm} \ref{alg:1}. Each byte is processed once in a single pass through the sequence, so the complexity is \(O(n)\). 

In addition to entropy, we also explore the use of \textbf{surprisal} as our information signal, noting that \tokname is compatible with any function mapping from the LM's logits to a scalar. We also consider a third constraint that combines the global constraint with the monotonic constraint, grouping contiguous bytes that meet either. This follows from the observation that the monotonic constraint can become unstable when the entropy or surprisal of a byte is close to zero. This can be seen from the small increase in surprisal from \example{l} to \example{e} causing \example{stable} to be split into two units in the example given in \cref{fig:example}. In such cases, the \green{combined constraint} joins segments below a low threshold \(\thres_g \) with monotonically decreasing segments, potentially preventing these unwanted segmentations.


The \red{global constraint} aims to capture highly predictable sequences. However, we find that it often results in splitting words into single byte tokens followed by the remainder of the word, and we find that the \yellow{monotonic constraint} is superior at recovering morphological and lexical units. This follows \citet{elman1990finding}'s observations that model uncertainty typically spike at such boundaries.

\begin{algorithm}[t]
\caption{\textbf{\tokname Tokenisation}}
\begin{algorithmic}[1]
\STATE \textbf{Input:} Byte sequence $X = \ch_1, \ch_2, \dots, \ch_n$, Byte-level entropy values $H(\ch_i)$
\STATE \textbf{Output:} Tokenized sequence $T$
\STATE Initialize $i \gets 2$ \COMMENT{Start of the byte sequence}
\STATE Initialize $T \gets []$ \COMMENT{Empty tokenized sequence}

\WHILE{$i \leq n$}
    \STATE $j \gets i$
    \WHILE{$j \leq n$ \textbf{and} $H(\ch_j) - H(\ch_{j-1}) < 0$}
        \STATE $j \gets j + 1$
    \ENDWHILE
    \STATE Extract segment $\ch_i, \ch_{i+1}, \dots, \ch_{j}$
    \STATE Append the segment to $T$
    \STATE $i \gets j$
\ENDWHILE

\STATE \textbf{Return:} $T$
\end{algorithmic}
\label{alg:1}
\end{algorithm}

\paragraph{Learning a vocabulary.} We propose three methods for using these constraints to learn a fixed-size vocabulary \(V\) from a training corpus $\dataset = \{\chvec_n\}_{n=1}^N$:
\begin{enumerate}
    \item \textbf{Frequency:} Using any of the three constraints, identify all unique subwords in the training corpus. Sort the subwords by frequency and use the top \(|V|\) as the tokeniser's vocabulary.
    \item \textbf{Incremental:} Using the \red{global constraint}, gradually increase $\thres_g$ until the desired vocabulary size is reached. To prevent rare subwords from being added to the vocabulary, a minimum frequency threshold $\thres_f$ can also be applied.
    \item \textbf{Seeding \bpe:} Using any of the three constraints, apply the \textbf{frequency cutoff} method to learn a portion $p\%$ of the final vocabulary, then apply \bpe to learn the rest of the vocabulary. 
\end{enumerate}

The frequency method is the most efficient, requiring only a single pass of the training dataset. However, for the \red{global constraint}, it requires pre-determining the global threshold $\thres_g$. The incremental method gets around this limitation by gradually increasing the threshold until the desired vocabulary size is reached. Unlike the other methods, this means that vocabularies with a larger \(|V|\) will not necessarily contain the vocabularies of tokenisers trained with a small \(|V|\). This is because higher thresholds lead to the \textbf{absorption} (or subsumption) of constituent subsequences. For instance, increasing the threshold in \cref{fig:example} would lead to $\example{nstable}$ replacing $\example{table}$ in the vocabulary.

In theory, the incremental method is also compatible with the \green{combined constraint}, but in practice, the number of subwords identified by the monotonic constraint exceeds most desired vocabulary sizes at the first pass, causing the algorithm to terminate immediately without the \red{global constraint} applying.

Finally, we theorise that the seeding method could provide a trade-off between \tokname and \bpe by first identifying predictable multi-byte units and then using \bpe to efficiently compress frequently co-occurring units. We note that setting $p=100\%$ is equivalent to the frequency method and that setting  $p=0\%$ is equivalent to just using \bpe to learn the vocabulary.

\paragraph{\texorpdfstring{\tokname \vs \bpe}{ByteSpan vs BPE}.}
Unlike \bpe, which merges tokens incrementally based on frequency, \tokname finds contiguous low-information segments. Our vocabulary-learning methods are flexible, achieving a target vocabulary size by either incrementally increasing the information threshold to include longer and less predictable sequences, or by trimming down a large set of discovered units according to frequency.

By only including the longest subsequences determined by the constraints used, we avoid intermediate merges, unlike \bpe. Since a sequence cannot be decomposed using a recursive \bpe-style inference procedure, we rely on \textbf{longest-prefix matching} as used by WordPiece (\wordpiece, \citealp{schuster-nakajima-2012-voice}). Notably, \tokname can also group beyond word boundaries and create \defn{superwords} (e.g., if the threshold were raised in \cref{fig:example}, \example{nstable and} could be added as a single token). Our implementation ensures that learned subwords align with \bpe pre-tokenisation constraints by preventing subwords from crossing pre-tokenisation boundaries (the vertical lines in \cref{fig:example}).

\paragraph{\texorpdfstring{\tokname \vs Patching}{ByteSpan vs Patching}.}
\tokname retains the information-driven approach of patching while preserving the computational benefits of keeping tokenisation separate from language modelling. In particular, it eliminates the additional complexity of batching in patch-based methods (e.g., where patch boundaries do not align across sequences). Our method only requires computing the entropy (or surprisal) of bytes once in order to learn a fixed \(\mathcal{V}\) before training, which can then be used by standard LMs with any corpus, whereas the patching method of \citet{pagnoni2024byte} requires a byte-level LM to compute the entropy of every byte seen during training. Unlike our method, patching does not create a fixed size vocabulary, as patches are dynamically created during training.

\section{Experimental Setup}
\label{sec:exp_setup}

We explore our information-driven tokenisation approach in both English and multilingual settings. Below, we briefly describe the experimental setup. We provide additional implementation details in \cref{app:implementation_details}.

\paragraph{Data.} 
We train the English tokenisers on a sample of the \fineweb dataset\footnote{\href{https://huggingface.co/datasets/HuggingFaceFW/fineweb-edu}{\myemph{huggingface.co/datasets/HuggingFaceFW/fineweb-edu}}.} \citep{penedo-etal-2024-fineweb}.
For the multilingual tokenisers, we use a balanced \integer{25}-language sample from the \commoncorpus \footnote{\href{https://huggingface.co/datasets/PleIAs/common_corpus}{\myemph{huggingface.co/datasets/PleIAs/common_corpus}}.} \citep{common_corpus}, with languages selected for morphological diversity. We convert each corpus into bytes using the Huggingface \texttt{ByteLevel} pre-tokenizer and split each corpus into three equal subsets: one to train the byte-level model, one to train the tokenizers, and one for evaluation. The \fineweb subsets contain approximately \q{500}{\million} bytes and the \commoncorpus subsets contain approximately \q{250}{\million} bytes (\q{10}{\million} per language).

\paragraph{Byte-level Model.} 
The byte-level LM is based on the Llama-2 architecture \citep{touvron2023llama} and we use it to collect the contextual surprisal and entropy for each byte in our corpora. The byte-level statistics only need to be collected once and can be reused for each tokeniser setup.

\paragraph{Tokeniser Setup.} 
For our English tokenisers, we train a suite of tokenisers across three vocabulary sizes; \q{16}{\thousand}, \q{32}{\thousand} and \q{64}{\thousand}. For the multilingual setting, we use a vocabulary size of \q{128}{\thousand}. For each vocabulary size, we train tokenisers using our three constraints and our two measures (entropy and surprisal). For the \red{global constraint} we use the incremental method for learning a vocabulary with a minimum frequency threshold $\thres_f = 20$. For the \yellow{monotonic constraint}, we use the frequency method and the seeding method with $p=50\%$. We use the same methods for the \green{combined constraint} and set the global threshold $\thres_g$ to be the 30th-percentile entropy (or surprisal) in the data.

As baselines, we train \bpe tokenisers for each vocabulary size and corpus. We also train \bpe tokenisers that use WP-style inference to match the inference procedure of our tokenisers, since inference can have a significant impact on intrinsic tokeniser evaluation \citep{uzan-etal-2024-greed}. We label these tokenisers \bpewp.\footnote{The WordPiece tokeniser trainers on Huggingface actually use \bpe, not the WordPiece objective, to learn their vocabularies.}

Our tokenisers (and \bpewp) are initialised with three copies of every byte in their vocabularies. In addition to each byte, these are the byte with the WordPiece \textbf{continuation prefix} \texttt{\#\#} (used by the inference method to distinguish pre-token-internal tokens from pre-word-initial tokens) and the byte with the start-of-word prefix (used by the \texttt{ByteLevel} pre-tokeniser to indicate whitespace). This slightly wastes vocabulary space compared to \bpe, as 768 base tokens are required instead of 512, but is a consequence of the complexities of combining tokeniser modules.

\begin{table*}[h]
    \centering
    \caption{Intrinsic evaluation results comparing \bpe and \bpewp to our \tokname tokenisers using surprisal across three vocabulary sizes. Scores are given to three significant figures, with the best score for each vocabulary size marked in \textbf{bold}.}
    \label{tab:englishresults}
    \vskip 0.15in
    \small
    \begin{sc}
    \begin{tabular}{cccccccc}
        \toprule
        Vocab Size & Tokenizer & Constraint & \makecell{Learning \\ Method} & \makecell{Morph. \\ Alignment} & \makecell{Cognitive \\ Plausibility} & Fertility & \makecell{Renyi \\ Efficiency} \\
        \midrule
        \multirow{7}{*}{\q{16}{\thousand}} & BPE & - & - & .694 & \textbf{.302} & 1.21 & .468 \\ 
         & BPE-WP & - & - & .834 & .297 & \textbf{1.19} & .472 \\ 
         & ByteSpan & \red{Global} & Increment & \textbf{.899} & .146 & 1.90 & .470 \\ 
         & ByteSpan & \yellow{Monotonic} & Frequency & .885 & .254 & 1.39 & \textbf{ .483} \\ 
         & ByteSpan & \yellow{Monotonic} & Seeding & .862 & .272 & 1.22 & .476 \\
         & ByteSpan & \green{Combined} & Frequency & .890 & .268 & 1.29 & .477 \\ 
         & ByteSpan & \green{Combined} & Seeding & .867 & .279 & 1.21 & .474 \\
        \midrule
        \multirow{7}{*}{\q{32}{\thousand}} & BPE & - & - & .648 & \textbf{.344} & 1.13 & .427 \\
         & BPE-WP & - & - & .821 & .337 & \textbf{ 1.11} & .431 \\
         & ByteSpan  & \red{Global} & Increment      & \textbf{.890} & .192 & 1.46 & \textbf{.466} \\
         & ByteSpan  & \yellow{Monotonic} & Frequency   & .843 & .277 & 1.31 & .446 \\
         & ByteSpan  & \yellow{Monotonic} & Seeding & .862 & .314 & 1.13 & .433 \\
         & ByteSpan  & \green{Combined} & Frequency    & .865 & .295 & 1.20 & .438 \\
         & ByteSpan  & \green{Combined} & Seeding  & .860 & .318 & 1.12 & .432 \\
        \midrule
        \multirow{7}{*}{\q{64}{\thousand}} & BPE & - & - & .609 & \textbf{.362} & 1.09 & .395 \\ 
         & BPE-WP & - & - & .773 & .358 & \textbf{1.06} & .399 \\ 
         & ByteSpan & \red{Global} & Increment & \textbf{.865} & .258 & 1.18 & \textbf{.421} \\ 
         & ByteSpan & \yellow{Monotonic} & Frequency & .816 & .285 & 1.25 & .416 \\ 
         & ByteSpan & \yellow{Monotonic} & Seeding & .809 & .339 & 1.08 & .410 \\ 
         & ByteSpan & \green{Combined} & Frequency & .833 & .302 & 1.14 & .409 \\ 
         & ByteSpan & \green{Combined} & Seeding & .808 & .344 & 1.07 & .400 \\ 
         \bottomrule
    \end{tabular}
    \end{sc}
    \vskip -0.1in
\end{table*}

\paragraph{Evaluation.} 
We use the intrinsic evaluation benchmark proposed in \citet{uzan-etal-2024-greed}. It consists of four information measures: Morphological Alignment, Cognitive Plausibility, \renyi Efficiency, and Fertility:

\begin{itemize}[leftmargin=*]
    \item \textbf{Morphological alignment} The alignment of tokenisers with gold-standard morphological segmentations of words. The benchmark compares tokeniser alignment to morphological annotations from seven resources and returns a macro-averaged $F_1$ score. The morphological data comes from LADEC \citep{gagne2019ladec}, MorphoLex \citep{sanchez2018morpholex}, MorphyNet \citep{batsuren-etal-2021-morphynet}, and DagoBert \citep{hofmann-etal-2020-dagobert}. \citet{uzan-etal-2024-greed} further augment these datasets with morpheme segmentation data \citep{batsuren-etal-2022-unimorph}, novel blend structure detection data \citep{pinter-etal-2021-will}, and compound separation data \citep{minixhofer-etal-2023-compoundpiece}.
    \item \textbf{Cognitive plausibility} A benchmark from \citet {beinborn-pinter-2023-analyzing} who found that human reaction time and accuracy from a lexical decision task correlated negatively with the number of tokens produced by a tokeniser for words, and correlated positively for nonwords. The score consists of an average from correlation scores across the four conditions (words/nonwords, accuracy/reaction time) with a higher score indicating increased `cognitive plausibility'.
    \item \textbf{\renyi efficiency} A measure that penalises vocabularies consisting of many low-frequency or high-frequency tokens. It captures efficient channel usage and was found to correlate highly with \textsc{Bleu} scores \citep{zouhar-etal-2023-tokenization}.
    \item \textbf{Fertility} The average number of subwords produced per tokenised word, used by \citet{acs2019exploring} to compare compression efficiency across languages for a multilingual tokeniser. A score of \integer{1} indicates perfect compression; every word in the evaluation set exists in the tokeniser's vocabulary.
\end{itemize}

This setup allows us to evaluate tokenisers without the time- and resource-intensive process of training and evaluating LMs on downstream tasks. 

\citet{uzan-etal-2024-greed} use \minipile \citep{kaddour2023minipile} to compute \renyi efficiency and fertility in order to match the dataset they used to train their tokenisers. To match our tokenisers, we compute these measures using the third subsets of our corpora (\fineweb for English, \commoncorpus for the multilingual tokenisers).

For our multilingual tokenisers, we only report \renyi efficiency and fertility using \commoncorpus, as the intrinsic evaluation benchmark for the other metrics only support English.

\section{Results}
\label{sec:results}

\begin{figure*}[!t]
    \centering
    \includegraphics[width=0.98\linewidth]{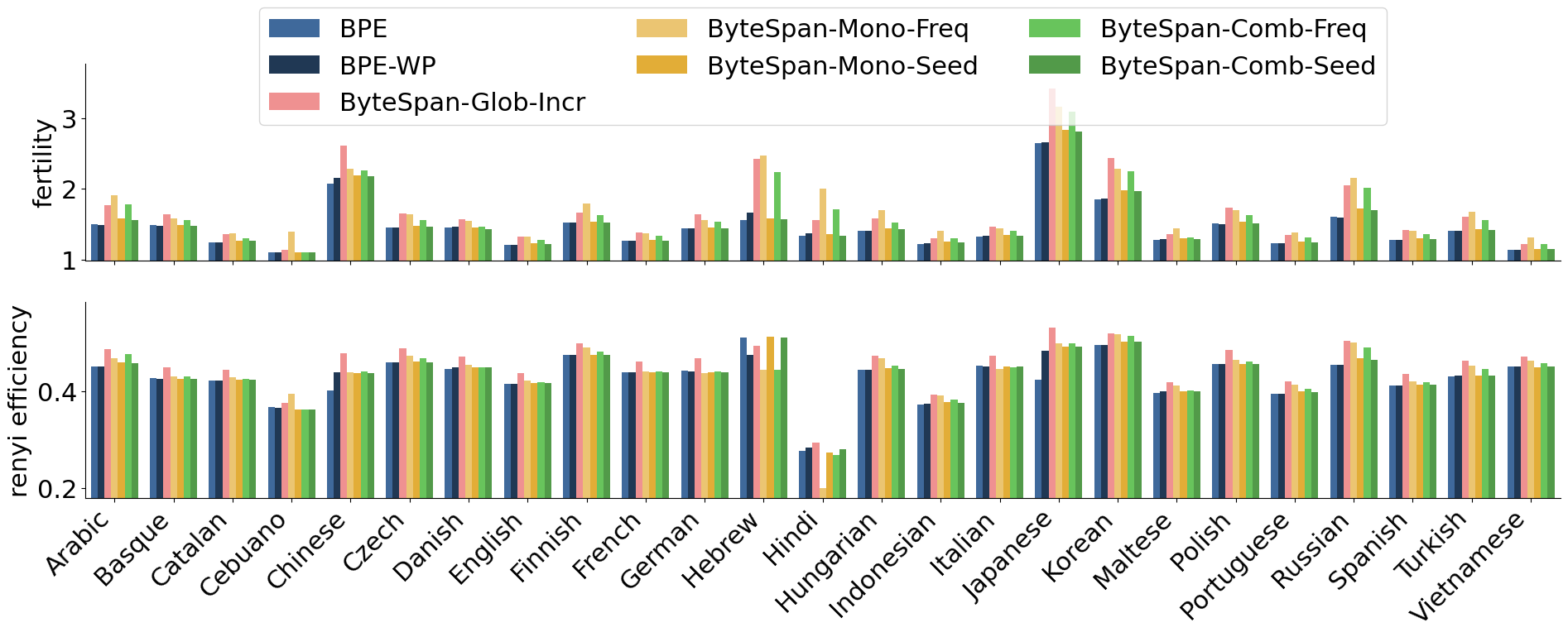}
    \caption{Fertility and r\'enyi efficiency for each language in our \commoncorpus evaluation subset, comparing multilingual \bpe to our multilingual \tokname tokenisers using surprisal with a vocabulary size of \q{128}{\thousand}.}
    \label{fig:multilingual}
\end{figure*}

We provide the results for our English tokenisers in \cref{tab:englishresults} and the results for our multilingual tokenisers in \cref{fig:multilingual}. We only include the \tokname tokenisers using \textbf{surprisal} as the information signal --- this is because for almost every measure, the equivalent tokenizer using \textbf{entropy} achieves almost identical results (the results with entropy are provided in \cref{app:additional_results}). When comparing vocabularies, we find a very high overlap --- ranging from \q{85.7}{\percent} to \q{98.8}{\percent} --- suggesting that both measures of information identify similar byte spans. We thus only report the \textbf{surprisal} scores in this section. Below, we summarise our key findings.

\paragraph{Linguistic and cognitive alignment.}

We find that our English \tokname tokenisers achieve higher morphological alignment scores than \bpe and \bpewp for each vocabulary size but achieve lower scores on the cognitive plausibility metric. This indicates that morphological units in text can be successfully extracted using the surprisal from a byte-level model but that the number of splits may not correlate well with human performance in a lexical decision task. Our results for \bpe and \bpewp mirror those of \citet{uzan-etal-2024-greed}, who found that using the vocabulary from \bpe but applying the WordPiece inference strategy led to higher morphological alignment but lower cognitive plausibility. In general, their results suggest a trade-off between these two measures, which we also observe with our tokenisers. It is unclear which score is more desirable, although morphological alignment has long been hypothesised to be important for downstream use-cases (see e.g. \citet{gow-smith-etal-2022-improving}).

When comparing between our proposed constraints and learning methods, the \red{global constraint} achieves the highest morphological alignment score for all three vocabulary sizes (and due to the apparent trade-off, also the lowest cognitive plausibility score). We found this result to contradict qualitative analysis of the tokenisers; in many cases, such as the example phrase in \cref{fig:example}, the \red{global constraint} seems to segment the first few letters of a word individually and then the rest of the word as one token, since surprisal tends to fall towards the end of a word. This does not seem to align with English morphology and in most examples, the \yellow{monotonic} or \green{combined} constraints seem to produce more morphological segmentations. Upon investigation of how the morphological alignment score is calculated in the intrinsic benchmark, we found that it skips words \textbf{unless all items in the gold segmentation of that word are contained in the vocabulary of the tokeniser}. For example, if a tokenizer's vocabulary does not contain the tokens \(\example{ramp}, \example{ant}, \example{ly}\) then the word \example{rampantly} is skipped. This can skew the score if a tokeniser's vocabulary does not contain many valid morphemes or stems, making comparison between tokenisers with different vocabularies difficult.

\begin{figure}[!t]
    \centering
    \includegraphics[width=0.95\linewidth]{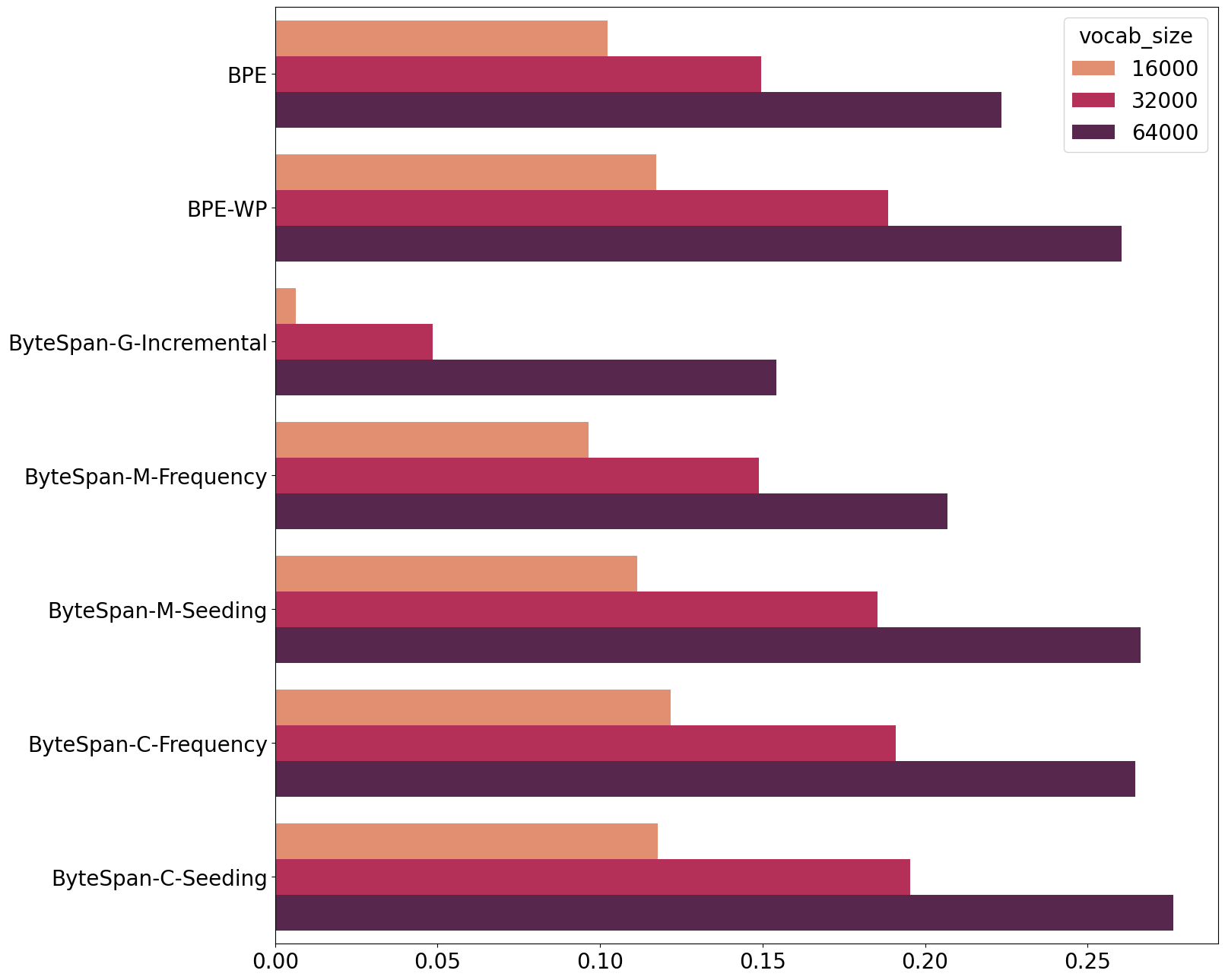}
    \caption{Morphological coverage of \bpe, \bpewp and our \tokname tokenisers using surprisal.}
    \label{fig:coverage}
\end{figure}

In order to explore this further, we define the \textbf{morphological coverage} of a tokeniser as the percentage of words in the morphological data where all gold segments exist in the tokeniser's vocabulary (i.e, the percentage of the words used to calculate the alignment score for each tokeniser). We plot the coverage in \cref{fig:coverage}. Indeed, the tokenisers using the \red{global constraint} have much lower coverage, suggesting that the high alignment score is not comparable to the other tokenisers. This analysis reveals that our other tokenisers have similar coverage to \bpe and \bpewp and still achieve higher morphological alignment, suggesting that they do produce a more linguistically motivated segmentation, with the \green{combined constraint} tokenisers achieving both high coverage and high alignment. 

\paragraph{Token distribution statistics.}

Besides those using the \red{global constraint}, our \tokname tokeniser lead to higher \renyi efficiency scores than \bpe and \bpewp and very similar fertility scores across vocabulary sizes. This indicates that \tokname leads to good compression while ensuring that the vocabulary space does not contain too many high-frequency and low-frequency tokens. Comparing between the incremental method and the seeding method for using \tokname to learn a vocabulary, we find that the seeding method improves fertility at the cost of \renyi efficiency, resulting in scores very similar to \bpewp. The fact that these tokenisers have similar token distribution statistics to \bpewp but maintain a higher morphological alignment score suggests that by using \tokname to learn an initial vocabulary that is then supplemented by \bpe, the resulting vocabulary contains more morphologically-aligned tokens without sacrificing compression.

\begin{figure}[!t]
    \centering
    \includegraphics[width=0.95\linewidth]{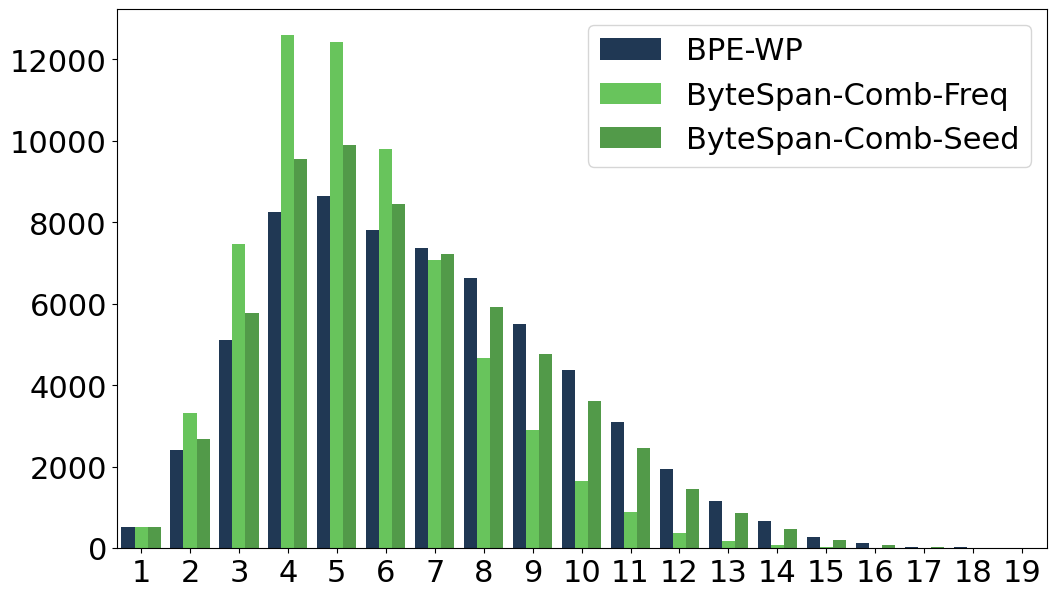}
    \caption{Distribution of token lengths comparing \bpewp to our \tokname tokenisers using surprisal and the \green{combined constraint} with either the frequency method or the seeding method to learn the vocabulary. Vocabulary size is \q{64}{\thousand}.}
    \label{fig:distribution}
\end{figure}

To investigate this further, we examine the length of each token in the vocabularies of the \bpewp tokeniser and our two tokenisers with the \green{combined constraint} for the largest vocabulary size, shown in \cref{fig:distribution}. For all three tokenisers, the most common token lengths are \integer{4} and \integer{5}, but the frequency method leads to a tighter distribution around these lengths compared to \bpewp. The seeding method provides a balance by allowing \bpe to merge commonly occurring sequences identified by \tokname, creating a distribution that more closely resembles the long tail of the \bpewp vocabulary.

In general, the \red{global constraint} does not lead to good compression. This is because, as observed in \cref{fig:example}, the first letters of words are highly unpredictable and so tend to be initially segmented at the character-level, even if long suffixes are compressed. The fact that this method learns long suffixes is also at odds with the longest-\textbf{prefix} inference method that we use. For example, for the largest tokeniser trained using this constraint, the token \example{bonization} is learned but \example{carbonization} is tokenised as \(\{\example{carbon},\example{ization}\}\) because the token \example{carbon} also exists in the vocabulary. The \yellow{monotonic} and \green{combined} constraints seem to learn units across words so are not as negatively affected by this inference method.

\paragraph{Multilingual evaluation.}

The fertility and \renyi efficiency scores for each tokeniser across the \integer{25} languages in our training corpus are given in \cref{fig:multilingual}. Note that the fertility scores are higher for Chinese, Japanese and Korean because these languages are not delimited by whitespace in \commoncorpus, so pre-tokens consist of entire phrases. 

Mirroring the English results, the \tokname tokenisers score achieve higher \renyi efficiency scores but lower fertility for most languages. Out of our tokenisers, the lowest fertility scores are achieved by the \yellow{monotonic} or \green{combined} constraints using the seeding method, whereas the incremental method and frequency method produce higher fertility scores. This difference is particularly pronounced for the languages in our corpus with unique writing systems; Arabic, Chinese, Hebrew, Hindi, Japanese, Korean and Russian. It is possible that since the frequency method adds the top $|V|$ most frequent byte spans to the vocabulary, this will naturally bias towards orthographies shared by most of the corpus (in this case, subwords containing Latin characters). Similarly, as the incremental method gradually raises the global threshold $\thres_g$, if the byte-level LM struggles to predict rarer orthographies due to occurring less frequently in the data, the threshold may not add as many subwords from those languages. In the case of the frequency method, allowing \bpe to learn the remaining $50\%$ of the vocabulary seems to be an effective strategy, but \bpe is also implicitly biased by frequency.

\begin{figure*}[!t]
    \centering
    \includegraphics[width=0.9\linewidth]{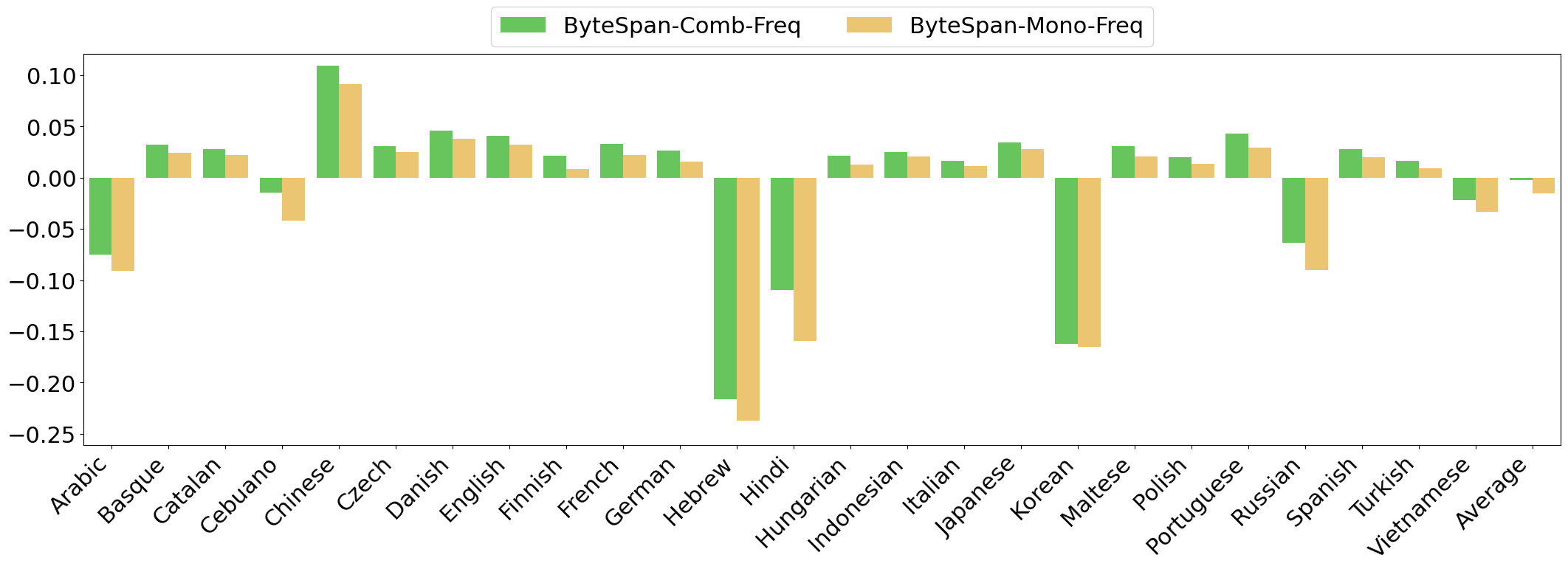}
    \caption{Increase in fertility for each language when balancing added tokens across languages when training our multilingual \tokname tokenisers with the frequency method. The tokenisers use surprisal as the byte-level measure and the vocabulary size is \q{128}{\thousand}. A decrease in fertility indicates better compression.}
    \label{fig:difference}
\end{figure*}

We consider one possible approach to this problem. For the frequency method, instead of selecting the top $|V|$ most frequent discovered subwords across the whole dataset, we can adjust the method to select the top $\frac{|V|}{L}$ most frequent discovered subwords for each language\footnote{Since subwords can appear in the frequency lists of multiple languages, we add tokens round-robin until the desired vocabulary size is met, which in practice can assign more than $\frac{|V|}{L}$ tokens from the languages with rarer orthographies.} (where $L$ is the number of languages). This guarantees that the rarer orthographies are assigned a dedicated portion of the final vocabulary. We plot the effect of this adjustment on fertility in \cref{fig:difference}. This approach seems to lead to the desired outcome; fertility decreases for most of the languages with unique orthographies (all but Chinese). By dedicating a portion of the vocabulary to these languages, the fertility does slightly rise for all Latin-script languages, but the average across languages is not affected (even slightly decreasing).

\section{Discussion}
\label{sec:discussion}

The proposed method of training a tokeniser based on information extracted from a byte-level LM achieves a balance between dynamic patching approaches and the computational benefits of a fixed vocabulary size. Through intrinsic evaluation in English, we found this approach to improve morphological alignment and \renyi efficiency compared to \bpe and \bpewp while retaining similar levels of compression. Whereas \citet{pagnoni2024byte} only used entropy in their study, we found surprisal to be an effective informative signal for grouping predictable bytes. As surprisal is cheaper to compute than entropy, this could lead to efficiency gains for dynamic patching approaches that require information to be calculated at every byte during training. Further work could explore information-based tokenisation with alternative information signals, such as the probability of whitespace tokens, mimicking connectionist models of word segmentation models that rely on utterance boundary prediction \citep{christiansen1998learning}.

In our multilingual evaluation, we found that the information-based approach struggles with languages whose writing systems are less represented in the data, but that using our method to seed an initial vocabulary before applying \bpe addresses a gap in compression for these languages. We note that our frequency method for learning a vocabulary may implicitly lead to the vocabulary being biased towards more the more frequent orthographies found in the training data, but that by adjusting our method to force the same number of subwords to be added for each language, we could improve fertility for languages with unique orthographies. In general, our method still relies on grouping contiguous bytes based on the predictions from a byte-level LM, which may not identify useful subwords for non-concatenative languages or right-to-left languages like Arabic and Hebrew. By using bytes as our fall-back representation, we also perpetuate the encoding bias of UTF-8, which on average assigns more bytes per character for non-Latin-script languages. Future work should incorporate more balanced byte-level schemes, such as MYTE, a morphologically-driven byte encoding \citep{limisiewicz-etal-2024-myte}.

Finally, although \tokname is parameter-free for certain combinations of constraints and vocabulary-learning methods, the \green{combined constraint} and \textbf{seeding method} both use hyper-parameters which we have not thoroughly explored here (we set $p=50\%$ for seeding method and $\thres_g$ to the 30th-percentile for the \green{combined constraint}). These should be explored further in future work. Future work could also explore the use of an \textbf{approximate} monotonic constraint (using $H(\ch_t) - H(\ch_{t-1}) < \thres_m$ instead of $H(\ch_t) - H(\ch_{t-1}) < 0$). This was proposed by \citet{pagnoni2024byte} and could provide an alternative mechanism of dealing with the instability of the monotonic constraint at near-zero values for surprisal and entropy.

\section{Conclusion}

We present \tokname, a novel method for learning a subword vocabulary using the predictions of a byte-level LM. By grouping contiguous sequences of predictable bytes using one of three constraints, we find that \tokname tokenisers have efficient vocabularies with a higher morphological alignment than \bpe and \bpewp on English evaluation sets, with the \yellow{monotonic constraint} generally being more effective than the \red{global constraint}. In the multilingual setting, \tokname tokenisers result in similar compression rates to \bpe but some methods struggle to compress languages with unique orthographies. We hypothesise that this could be due to our frequency-based vocabulary-learning method and find that balancing the vocabulary by language counteracts this effect. In general, \tokname provides a novel method for learning subwords with parallels to word segmentation and patching, all while keeping the benefits of learning a fixed-size vocabulary for efficient language modelling. This could feed into explorations of information in lexical unit extraction that may improve future static tokenisers and dynamic patching approaches.

\section*{Limitations}

In this study we propose novel methods for learning a subword vocabulary but only use one inference method for applying the tokenisers; the longest-prefix method of WordPiece. Future work could explore the use our learned vocabularies with alternative inference methods, such as longest-suffix matching \citep{jacobs2022lost}.

The experiments reported in the paper only utilise one type of Transformer models, although preliminary work utilised a 5-gram model. Further work might explore the scaling properties of the byte-level model. Our evaluation largely focused on English, although we do explore a multilingual setting. In our multilingual setting we are restricted to token distribution statistics due to limited evaluation resources available for these other languages, although there are individual pipelines for individual languages (e.g., \citet{gazit2025splinteringnonconcatenativelanguagesbetter} uses lexical decision data to evaluate Hebrew tokenisers).

Finally, in this study, we have focused on intrinsic evaluation of tokenisation methods, in part due to the computational cost of extrinsic evaluation. The intrinsic evaluation benchmark scores are designed to allude to potential gains in language modelling capability, but ideally this should be established by pre-training separate language models with each tokeniser and evaluating the pre-trained models using perplexity on held-out data or grammatical benchmarks such as \blimp \citep{warstadt2020blimp}.

\section*{Impact Statement}




This paper presents work whose goal is to advance the field of Machine Learning. There are many potential societal consequences of our work, none which we feel must be specifically highlighted here.

\bibliography{biblio}
\bibliographystyle{icml2025}

\appendix
\onecolumn

\section{Implementation Details}
\label{app:implementation_details}

We implement all experiments using the PyTorch framework \citep{paszke-etal-2019-pytorch} and implement variants of the Llama architecture with components implemented in the \myemph{transformers} library \citep{wolf-etal-2020-transformers}. Tokenisers are implemented using modules from the \myemph{tokenizers} library.\footnote{\href{https://github.com/huggingface/tokenizers}{\myemph{github.com/huggingface/tokenizers}}.}




\paragraph{Byte-Level Model Training.}
We train a small byte-level LMs on our subsets of the \fineweb and \commoncorpus datasets. 
We use the Llama 2 architecture with \integer{24} attention heads, \integer{6} layers, hidden size of \integer{768}, and tied input--output embeddings, totalling 57M parameters.
We use AdamW \citep{adam,loshchilov2018decoupled} as our optimiser, with learning rate \snum{6e-4}, parameters $\beta_1\mathop{=}0.9$, $\beta_2\mathop{=}0.95$, $\epsilon\mathop{=}\snum{1e-8}$, and weight decay set to \float{0.1}.
We use the warm-up-stable-decay schedule \citep{zhai2022scaling}, where after warm-up, the learning rate stays constant for most training and decreases briefly during the cool-down. 
Unlike the cosine schedule, this approach does not need a pre-specified compute budget, facilitating continued pre-training and enhancing our artifact's value for the community. It is also more effective for small language models \citep{hu2024minicpm, wen2024understanding}.
During training, we set the context size to \integer{2048} tokens and batch size \integer{128}, clip the norm of the gradients to \float{1}, and train for \q{50}{\thousand} steps, saving checkpoints every \q{2}{\thousand} steps.

\paragraph{Byte-Level Model Predictions.}

We use our trained models to extract per-byte entropy and surprisal on a different subsets of \fineweb and \commoncorpus to prevent overfitting. We use a context size of \integer{2048} and shift the dataset along by strides of \integer{512}. This ensures that all byte-level predictions have at least \integer{1536} tokens of context. We then use the logits at each byte to calculate surprisal and entropy. These predictions are stored as a split of each dataset in Huggingface to facilitate training tokenisers using our method.

\paragraph{Hardware Details.}
We use a server with one \myemph{NVIDIA A100 80GB PCIe}, \integer{32} CPUs, and \integer{32} GB of RAM for all experiments. Below, we report a subset of the output of the \myemph{lscpu} command:

\begin{tcolorbox}[left=5pt,right=5pt,top=5pt,bottom=5pt]
    \small
    \begin{verbatim}
Architecture:        x86_64
CPU op-mode(s):      32-bit, 64-bit
Address sizes:       46 bits physical, 
                     48 bits virtual
Byte Order:          Little Endian
CPU(s):              32
On-line CPU(s) list: 0-31
Vendor ID:           GenuineIntel
Model name:          Intel(R) Xeon(R)
                     Silver 4210R CPU
                     @ 2.40GHz
CPU family:          6
Model:               85
Thread(s) per core:  1
Core(s) per socket:  1
Socket(s):           8
Stepping:            7
BogoMIPS:            4800.11
\end{verbatim}
\end{tcolorbox}

\paragraph{Reproducibility.}

We release all experimental artefacts as a collection on the Hugging Face Hub at \href{https://huggingface.co/InfoTokenizers}{huggingface.co/InfoTokenizers}: (i) the byte-level versions of the two datasets; (ii) the subsets of each dataset used to train the byte-level models, extract predictions and evaluate the resulting tokenisers; (iii) the byte-level surprisal and entropy for each dataset; (iv) the \bpe, \bpewp and \tokname tokenisers used in our experiments; (v) all model checkpoints. Our codebase is available at \href{https://github.com/codebyzeb/bytespantokenization}{github.com/codebyzeb/bytespantokenization}.\looseness=-1

\section{Additional Results}
\label{app:additional_results}

\begin{figure*}[!t]
    \centering
    \includegraphics[width=0.98\linewidth]{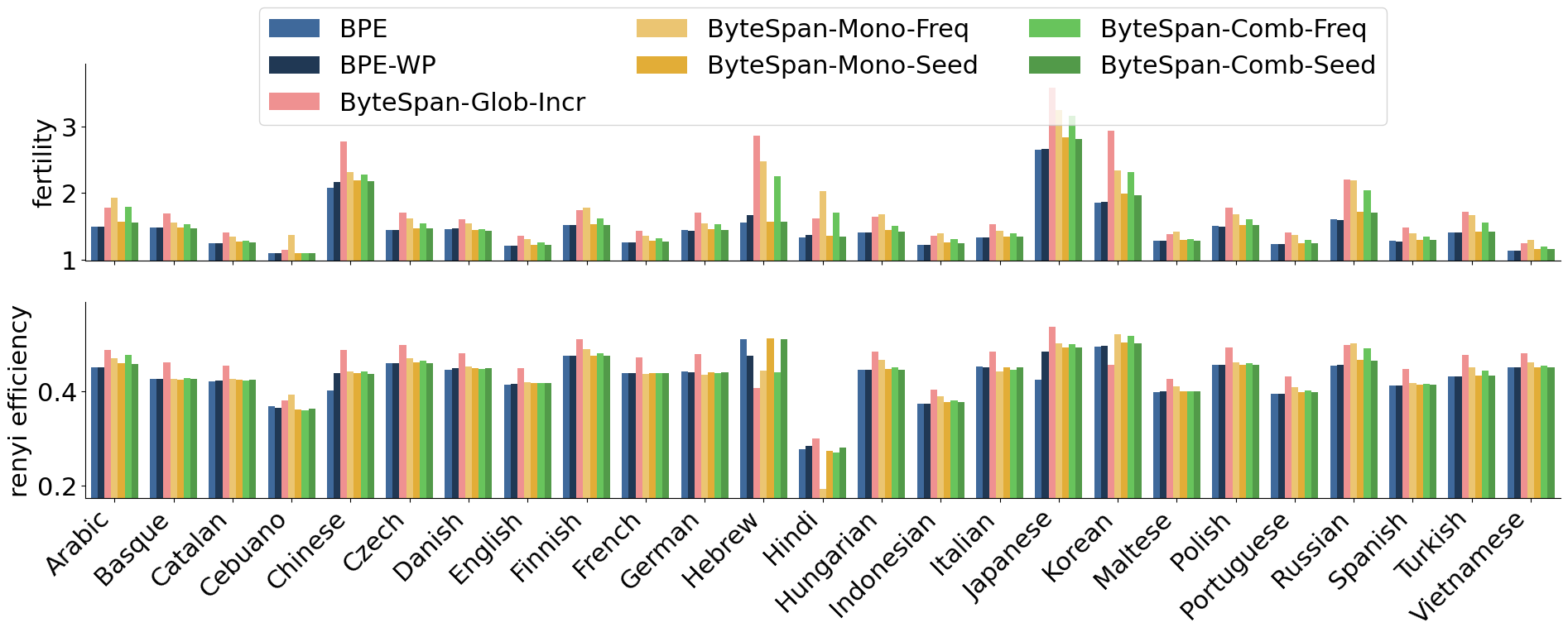}
    \caption{Fertility and r\'enyi efficiency for each language in our \commoncorpus evaluation subset, comparing multilingual \bpe to our multilingual \tokname tokenisers using entropy with a vocabulary size of \q{128}{\thousand}.}
    \label{fig:multilingualentropy}
\end{figure*}

For brevity, in \cref{sec:results} we only included results for the \tokname tokenisers using surprisal. Here, we provide the full results for our English tokenisers in \cref{tab:fullenglishresults}, and the results for our multilingual tokenisers using entropy in \cref{fig:multilingualentropy}. As discussed in \cref{sec:results}, we observe very little difference between the two measures.

\begin{table*}[h]
    \centering
    \caption{Intrinsic evaluation results comparing \bpe and \bpewp to our \tokname tokenisers using surprisal (ByteSpan-S) and entropy (ByteSpan-E) across three vocabulary sizes. Scores are given to three significant figures, with the best score for each vocabulary size marked in \textbf{bold}.}
    \label{tab:fullenglishresults}
    \vskip 0.15in
    \small
    \begin{sc}
    \begin{tabular}{cccccccc}
        \toprule
        Vocab Size & Tokenizer & Constraint & \makecell{Learning \\ Method} & \makecell{Morph. \\ Alignment} & \makecell{Cognitive \\ Plausibility} & Fertility & \makecell{Renyi \\ Efficiency} \\
        \midrule
        \multirow{12}{*}{\q{16}{\thousand}} & BPE  & - & - & .694 & \textbf{.302} & 1.21 & .468 \\ 
         & BPE-WP & - & - & .834 & .297 & \textbf{1.19} & .472 \\ 
         & ByteSpan-S & \red{Global} & Increment & \textbf{.899} & .146 & 1.90 & .470 \\ 
         & ByteSpan-E & \red{Global} & Increment & \textbf{.899} & .146 & 1.90 & .470 \\ 
         & ByteSpan-S  & \yellow{Monotonic} & Frequency & .885 & .254 & 1.39 & \textbf{ .483} \\ 
         & ByteSpan-E & \yellow{Monotonic} & Frequency & .882 & .271 & 1.38 &  .482 \\ 
         & ByteSpan-S & \yellow{Monotonic} & Seeding & .862 & .272 & 1.22 & .476 \\
         & ByteSpan-E  & \yellow{Monotonic} & Seeding & .857 & .273 & 1.22 & .476 \\
         & ByteSpan-S  & \green{Combined} & Frequency & .890 & .268 & 1.29 & .477 \\ 
         & ByteSpan-E & \green{Combined} & Frequency & .889 & .286 & 1.27 & .476 \\ 
         & ByteSpan-S  & \green{Combined} & Seeding & .867 & .279 & 1.21 & .474 \\
         & ByteSpan-E & \green{Combined} & Seeding & .866 & .281 & 1.21 & .474 \\
        \midrule
        \multirow{12}{*}{\q{32}{\thousand}} & BPE & - & - & .648 & \textbf{.344} & 1.13 & .427 \\
         & BPE-WP  & - & - & .821 & .337 & \textbf{ 1.11} & .431 \\
         & ByteSpan-S & \red{Global} & Increment      & \textbf{.890} & .192 & 1.46 & .466 \\
         & ByteSpan-E & \red{Global} & Increment      & \textbf{.890} & .193 & 1.48 & \textbf{.467} \\
         & ByteSpan-S  & \yellow{Monotonic} & Frequency   & .843 & .277 & 1.31 & .446 \\
         & ByteSpan-E & \yellow{Monotonic} & Frequency   & .837 & .291 & 1.30 & .445 \\
         & ByteSpan-S & \yellow{Monotonic} & Seeding & .862 & .314 & 1.13 & .433 \\
         & ByteSpan-E  & \yellow{Monotonic} & Seeding & .859 & .315 & 1.13 & .433 \\
         & ByteSpan-S & \green{Combined} & Frequency    & .865 & .295 & 1.20 & .438 \\
         & ByteSpan-E  & \green{Combined} & Frequency    & .854 & .308 & 1.19 & .438 \\
         & ByteSpan-S & \green{Combined} & Seeding  & .860 & .318 & 1.12 & .432 \\
         & ByteSpan-E  & \green{Combined} & Seeding  & .858 & .322 & 1.12 & .432 \\
        \midrule
        \multirow{12}{*}{\q{64}{\thousand}} & BPE & - & - & .609 & \textbf{.362} & 1.09 & .395 \\ 
         & BPE-WP & - & - & .773 & .358 & \textbf{1.06} & .399 \\ 
         & ByteSpan-S  & \red{Global} & Increment & \textbf{.865} & .258 & 1.18 & .421 \\ 
         & ByteSpan-E  & \red{Global} & Increment & \textbf{.865} & .271 & 1.20 & \textbf{.424} \\ 
         & ByteSpan-S & \yellow{Monotonic} & Frequency & .816 & .285 & 1.25 & .416 \\ 
         & ByteSpan-E  & \yellow{Monotonic} & Frequency & .806 & .293 & 1.24 & .416 \\ 
         & ByteSpan-S  & \yellow{Monotonic} & Seeding & .809 & .339 & 1.08 & .410 \\ 
         & ByteSpan-E  & \yellow{Monotonic} & Seeding & .805 & .340 & 1.08 & .416 \\ 
         & ByteSpan-S  & \green{Combined} & Frequency & .833 & .302 & 1.14 & .409 \\ 
         & ByteSpan-E  & \green{Combined} & Frequency & .822 & .307 & 1.13 & .408 \\ 
         & ByteSpan-S  & \green{Combined} & Seeding & .808 & .344 & 1.07 & .400 \\ 
         & ByteSpan-E & \green{Combined} & Seeding & .804 & .344 & 1.07 & .400 \\ 
         \bottomrule
    \end{tabular}
    \end{sc}
    \vskip -0.1in
\end{table*}

\end{document}